\documentclass[conference]{IEEEtran}
\IEEEoverridecommandlockouts
\usepackage{cite}
\usepackage{amsmath,amssymb,amsfonts}
\usepackage{algorithmic}
\usepackage{graphicx}
\usepackage{textcomp}
\usepackage{xcolor}
\usepackage{url}

\usepackage{booktabs,makecell, multirow, tabularx}
\usepackage{microtype}		
\usepackage{stfloats}
\usepackage{hyperref}
\hypersetup{
    colorlinks=true,
    linkcolor=blue,
    citecolor=green,
    urlcolor=magenta,
    filecolor=cyan,
}
\usepackage{cleveref} 		
\crefname{figure}{Fig.}{Figs.}
\crefname{table}{Table}{Tables}
\crefname{equation}{Eq.}{Eqs.}
\usepackage[dvipsnames,table]{xcolor} 

\def\BibTeX{{\rm B\kern-.05em{\sc i\kern-.025em b}\kern-.08emT\kern-.1667em\lower.7ex\hbox{E}\kern-.125emX}}

\begin{document}

\title{Risk-Routed Implicit Boundary Refinement for Robust Ultrasound Image Segmentation
	\thanks{* Corresponding author. This work was supported by the General Research Funds of the Research Grant Council of Hong Kong under Grant 15102222 and Grant 15102524.}
}

\author{
	\IEEEauthorblockN{
		Jingguo Qu\textsuperscript{1},
		Xinyang Han\textsuperscript{1},
		Xiang Wang\textsuperscript{1},
		Yuqi Yang\textsuperscript{1},
		Tonghuan Xiao\textsuperscript{1},
		Sheng Ning\textsuperscript{1},\\
		Jing Qin\textsuperscript{2},
		Ann Dorothy King\textsuperscript{3},
		Winnie Chiu-Wing Chu\textsuperscript{3},
		Jing Cai\textsuperscript{1},
		and Michael Ying\textsuperscript{1,*}
	}
	\IEEEauthorblockA{
		\textsuperscript{1}\textit{Department of Health Technology and Informatics, The Hong Kong Polytechnic University, Hong Kong, China}\\
		\textsuperscript{2}\textit{Centre for Smart Health and School of Nursing, The Hong Kong Polytechnic University, Hong Kong, China}\\
		\textsuperscript{3}\textit{Department of Imaging and Interventional Radiology, The Chinese University of Hong Kong, Hong Kong, China}\\
	}
}

\maketitle

\begin{abstract}
Accurate boundary delineation in medical ultrasound (US) images remains challenging because of speckle noise, low contrast, acoustic shadowing, and acquisition variability. We propose Risk-routed Implicit Boundary Refinement (RIBR), a compact framework that refines convolutional segmentation through spatially weighted implicit residuals. A coordinate-conditioned implicit branch uses regional features to predict logit corrections, and a learned gate controls their contribution using predictive uncertainty and local appearance. Geometry- and speckle-aware regularization provides contour supervision and discourages probability fluctuations away from boundaries. We compare RIBR with ten baselines on nine datasets covering lymph nodes, breast lesions, thyroid nodules, and prostate, including four external lymph-node and breast cohorts. Across three seeds, RIBR achieves the best overall dataset macro-average among the evaluated configurations, with $76.67{\pm}0.42\%$ Dice and a 95th-percentile Hausdorff distance (HD95) of $29.55{\pm}0.55$ pixels on the resized evaluation grid. It obtains the lowest HD95 on all nine datasets, with task-dependent Dice rankings. These results support controlled implicit residual refinement for boundary-sensitive US segmentation. Source code is available at \href{https://github.com/jinggqu/ribr}{GitHub}.
\end{abstract}


\section{Introduction}
Ultrasound (US) is widely used for the assessment of breast lesions, thyroid nodules, lymph nodes (LNs), and the prostate. Automated delineation can improve the consistency of quantitative assessment. Speckle noise, acoustic shadowing, weak margins, and acquisition variation make boundary localization difficult, even when the approximate target region is correctly identified.

U-Net~\cite{ronneberger2015unet} and its variants~\cite{zhou2018unetpp,oktay2018attention} combine semantic context with spatial detail. Automated configuration~\cite{isensee2021nnunet}, transformers~\cite{hatamizadeh2022unetr,hatamizadeh2022swinunetr,chen2021transunet,cao2022swinunet}, and state-space modeling~\cite{ruan2024vmunet} strengthen regional representations. US-focused approaches address appearance and annotation variability through adaptive attention in AAU-Net~\cite{chen2023aau}, semi-supervised learning in PH-Net~\cite{jiang2024phnet} and Switch~\cite{qu2026multiscale}, and pretraining or encoder adaptation~\cite{jiao2024usfm,qu2026adapting}. SABR-Net~\cite{chen2023sabr} combines shadow-aware learning with boundary and direction maps that resample predictions at shifted pixel locations.

Contour-sensitive objectives complement these architectural advances. Dice~\cite{milletari2016vnet} and focal loss~\cite{lin2017focal} supervise region predictions, while boundary~\cite{kervadec2021boundary} and Hausdorff-distance-related losses~\cite{karimi2020hausdorff} provide geometric supervision. Recovering fine contours also depends on how spatial detail is represented. Implicit neural representations (INRs) model this detail through coordinate-based prediction: SIREN~\cite{sitzmann2020siren} uses sinusoidal activations, and MetaSeg~\cite{vyas2025metaseg} predicts complete label fields through image-specific adaptation. In US images, weak anatomical margins and speckle create ambiguous local structure, motivating regional conditioning and spatial control of coordinate-based corrections.

We propose Risk-routed Implicit Boundary Refinement (RIBR) for spatially controlled residual correction of convolutional logits. Boundary-refinement implicit residuals (BRI) provide coordinate-conditioned corrections; risk-routed residual control (RRC) weights their contribution using uncertainty and local appearance; geometry- and speckle-aware boundary regularization (GSBR) supervises contours and discourages non-boundary oscillation. We evaluate RIBR against ten baselines on nine datasets, including four external LN and breast cohorts.

\section{Method}

\subsection{Overall Framework}
\begin{figure*}[t]
    \centering
    \includegraphics[width=0.85\textwidth]{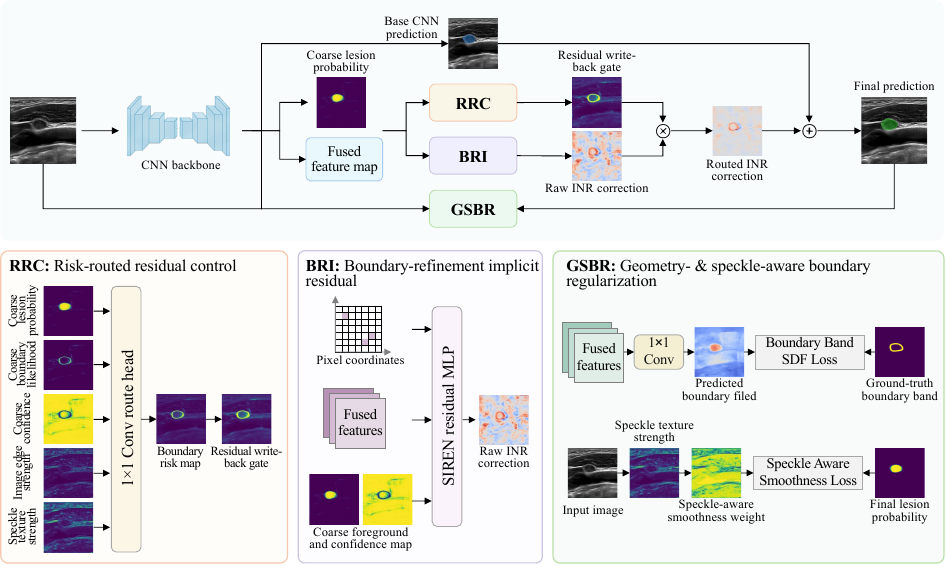}
    \caption{RIBR combines convolutional logits with gated implicit residuals. BRI supplies corrections, RRC controls their contribution, and GSBR provides training supervision.}
    \label{fig:ribr_overview}
\end{figure*}

As shown in~\Cref{fig:ribr_overview}, RIBR combines convolutional regional prediction with BRI, RRC, and GSBR. Given a single-channel image $I$, a compact U-Net-style encoder-decoder produces conditioning logits $\widetilde z_i$ and probability $p_i=\mathrm{softmax}(\widetilde z_i)_1$ at pixel $i$. A convolutional refinement path combines spatial and semantic predictions into base logits $z_i^0\in\mathbb{R}^2$. BRI predicts a two-class residual $\delta_i$, and RRC assigns a gate $\alpha_i\in[0,1]$. The final logits and foreground probability are
\begin{equation}
    z_i=z_i^0+\beta\alpha_i\delta_i,\qquad
    \widehat p_i=\mathrm{softmax}(z_i)_1,
    \label{eq:ribr_final}
\end{equation}
where $\beta$ scales the correction.

\subsection{Boundary-Refinement Implicit Residual}
BRI conditions logit corrections on spatial position and convolutional context using a two-hidden-layer SIREN:
\begin{equation}
    \delta_i=f_\theta([x_i,\rho_i,F_i,m_i,s_i,p_i,c_i]).
    \label{eq:bri}
\end{equation}
Here, $x_i\in[-1,1]^2$ is the image coordinate. The object-relative coordinate $\rho_i$ centers on the soft foreground centroid and scales each axis by its probability-weighted standard deviation, with a lower bound on scales and clipped coordinates. $F_i$ combines local decoder features, global anatomical and appearance context, the conditioning prediction, and fixed-grid depth and direction cues. Local intensity mean $m_i$ and standard deviation $s_i$ use $5 \times 5$ neighborhoods; $c_i=1-H(p_i)/\log2$ is confidence, with binary entropy $H$.

\subsection{Risk-Routed Residual Control}
RRC controls correction strength using uncertainty and local appearance. Its prior $b_i=1-2|p_i-0.5|$ peaks at equal foreground/background probability. A pointwise head $h_\phi$ uses $r_i=[b_i,p_i,c_i,m_i,s_i,g_i]$, where $g_i$ is normalized image-gradient magnitude, to compute
\begin{equation}
    \begin{aligned}
        q_i&=\mathrm{clip}(b_i+\lambda\tanh(h_\phi(\mathrm{sg}(r_i))),0,1),\\
        \alpha_i&=\sigma(\gamma(q_i-\tau)).
    \end{aligned}
    \label{eq:gate}
\end{equation}
Here, $\sigma$ is the sigmoid; $\tau$ and $\gamma$ set its threshold and slope. The adjustment $\lambda\tanh(\cdot)$ can raise or lower the prior by at most $\lambda$. Zero initialization of the head starts routing from the prior. Stop-gradient $\mathrm{sg}$ applies only to the head input, leaving the direct path through $b_i$ differentiable. BRI evaluates every pixel, and the gate scales its residual contribution.

\subsection{Boundary Regularization and Training}
GSBR adds contour supervision and local consistency through three terms. A narrow-band absolute-error loss $\mathcal{L}_{\mathrm{sdf}}$ supervises an auxiliary signed-distance head from $F_i$ using truncated, normalized inside-minus-outside Euclidean contour distances. Boundary-weighted cross entropy $\mathcal{L}_{\mathrm{bce}}$ uses weights $1+\kappa B_i$, normalized by their sum, where reference band $B$ is obtained by $5 \times 5$ dilation minus erosion. Speckle-aware smoothing penalizes probability variation away from this band:
\begin{equation}
    \mathcal{L}_{\mathrm{sp}}=\frac12\sum_{a\in\{x,y\}}
    \frac{\sum_{(i,j)\in\mathcal{N}_a}w_{ij}(\widehat p_i-\widehat p_j)^2}
    {\max(\sum_{(i,j)\in\mathcal{N}_a}w_{ij},1)},
    \label{eq:smoothness}
\end{equation}
Here, $\mathcal{N}_x$ and $\mathcal{N}_y$ contain horizontal and vertical neighbors. We set $w_{ij}=\min\{\sigma(-4s_i),\sigma(-4s_j)\}$ when both pixels lie outside the reference band, and zero otherwise. Higher intensity variation reduces smoothing. Reference masks are used only for training.

Focal and Dice losses supervise the final prediction; $\mathcal{L}_{\mathrm{coarse}}$ sums cross entropy and Dice on the conditioning prediction. In RRC variants, $\mathcal{L}_{\mathrm{res}}=\mathrm{mean}(\delta^2)$ penalizes the ungated residual over all pixels and classes in the batch. The objective is
\begin{equation}
    \begin{aligned}
        \mathcal{L}={}&\lambda_f\mathcal{L}_{\mathrm{focal}}+\lambda_d\mathcal{L}_{\mathrm{dice}}
        +\lambda_c\mathcal{L}_{\mathrm{coarse}}\\
        &+\lambda_b\mathcal{L}_{\mathrm{bce}}+\lambda_s\mathcal{L}_{\mathrm{sdf}}
        +\lambda_p\mathcal{L}_{\mathrm{sp}}+\lambda_r\mathcal{L}_{\mathrm{res}}.
    \end{aligned}
    \label{eq:objective}
\end{equation}
Ablations use the losses of their enabled modules, with $\delta_i=0$ when BRI is removed and $\alpha_i=1$ when RRC is removed. Inference takes the class argmax of $z$; the distance head supplies training supervision.

\section{Experiments}

\subsection{Datasets and Implementation}
\label{sec:implementation}
We evaluate nine US datasets with the splits in~\Cref{tab:datasets}. LN-INT contains retrospective cervical LN images from Prince of Wales Hospital, Hong Kong, annotated by experienced radiologists~\cite{qu2026multiscale}. LymphUS-C1/C2 are held-out cohorts from two external centers~\cite{mohammadi2026lymphus}. BUSI excludes normal images without lesion masks; Breast-USG and BUS-UCLM are external breast cohorts. External sets are excluded from training, validation, and hyperparameter selection and evaluated after LN-INT or BUSI checkpoint selection. DDTI, TN3K, and Prostate use their own partitions.

\begin{table}[htb]
    \caption{Dataset partitions in images. Ext. denotes held-out external test cohorts.}
    \label{tab:datasets}
    \centering
    \begin{tabular}{lrrrrr}
        \toprule
        Dataset                                & Train & Val. & Test & Ext. & Total \\
        \midrule
        LN-INT                                 & 1,018 & 127  & 128  & -   & 1,273 \\
        LymphUS-C1~\cite{mohammadi2026lymphus} & -    & -   & -   & 180  & 180   \\
        LymphUS-C2~\cite{mohammadi2026lymphus} & -    & -   & -   & 158  & 158   \\
        \midrule
        BUSI~\cite{al2020busi}                 & 517   & 64   & 66   & -   & 647   \\
        Breast-USG~\cite{pawlowska2024curated} & -    & -   & -   & 252  & 252   \\
        BUS-UCLM~\cite{vallez2025bus}          & -    & -   & -   & 174  & 174   \\
        \midrule
        DDTI~\cite{pedraza2015ddti}            & 509   & 63   & 65   & -   & 637   \\
        TN3K~\cite{gong2021tn3k}               & 2,265 & 614  & 614  & -   & 3,493 \\
        Prostate~\cite{jiang2024microsegnet}   & 1,241 & 690  & 690  & -   & 2,621 \\
        \bottomrule
    \end{tabular}
\end{table}

\begin{table*}[!b]
    \caption{Segmentation results and dataset macros, mean $\pm$ SD over three seeds. Dice is in percent; finite-only HD95 is in resized-grid pixels. Params are trainable parameters. \textbf{Best} and \underline{second-best} results are highlighted in bold and underlined.}
    \label{tab:results}
    \centering
    \begin{tabular}{lr|rr|rr|rr|rr}
        \toprule
        \multirow{2}{*}{\textbf{Method}}          & \multirow{2}{*}{\textbf{Params}} & Dice\%$\uparrow$                     & HD95$\downarrow$                         & Dice\%$\uparrow$                         & HD95$\downarrow$                                & Dice\%$\uparrow$                    & HD95$\downarrow$                    & Dice\%$\uparrow$                    & HD95$\downarrow$                    \\
        \cmidrule{3-10}
                                                  &                                  & \multicolumn{2}{c|}{\textbf{LN-INT}} & \multicolumn{2}{c|}{\textbf{LymphUS-C1}} & \multicolumn{2}{c|}{\textbf{LymphUS-C2}} & \multicolumn{2}{c}{\textbf{Lymph Node Average}}                                                                                                                                                         \\
        \midrule
        U-Net~\cite{ronneberger2015unet}          & 1.8~M                            & 73.95{\tiny\textpm3.06}              & 30.33{\tiny\textpm3.96}                  & 45.84{\tiny\textpm7.25}                  & 47.54{\tiny\textpm5.42}                         & 58.11{\tiny\textpm5.65}             & 54.70{\tiny\textpm5.43}             & 59.30{\tiny\textpm3.92}             & 44.19{\tiny\textpm2.81}             \\
        U-Net++~\cite{zhou2018unetpp}             & 2.3~M                            & 74.00{\tiny\textpm1.99}              & 32.67{\tiny\textpm2.93}                  & 45.59{\tiny\textpm4.19}                  & 56.93{\tiny\textpm6.05}                         & 61.07{\tiny\textpm4.34}             & 55.89{\tiny\textpm3.98}             & 60.22{\tiny\textpm2.14}             & 48.50{\tiny\textpm4.18}             \\
        Attention U-Net~\cite{oktay2018attention} & 2.0~M                            & 73.63{\tiny\textpm3.73}              & 31.72{\tiny\textpm3.30}                  & 42.79{\tiny\textpm3.92}                  & 53.89{\tiny\textpm2.52}                         & 58.52{\tiny\textpm5.68}             & 52.13{\tiny\textpm3.37}             & 58.31{\tiny\textpm4.13}             & 45.91{\tiny\textpm3.06}             \\
        MetaSeg~\cite{vyas2025metaseg}            & 0.1~M                            & 45.59{\tiny\textpm1.53}              & 51.88{\tiny\textpm2.33}                  & 34.46{\tiny\textpm2.15}                  & 48.02{\tiny\textpm1.38}                         & 44.88{\tiny\textpm0.18}             & 54.57{\tiny\textpm1.29}             & 41.64{\tiny\textpm1.08}             & 51.49{\tiny\textpm0.85}             \\
        DynUNet~\cite{isensee2021nnunet}          & 2.0~M                            & 72.38{\tiny\textpm2.67}              & 36.46{\tiny\textpm4.46}                  & 51.20{\tiny\textpm5.44}                  & 61.34{\tiny\textpm5.16}                         & 57.93{\tiny\textpm4.02}             & 57.67{\tiny\textpm2.45}             & 60.50{\tiny\textpm4.00}             & 51.82{\tiny\textpm3.24}             \\
        UNETR~\cite{hatamizadeh2022unetr}         & 13.9~M                           & 67.67{\tiny\textpm2.13}              & 40.60{\tiny\textpm2.54}                  & 59.86{\tiny\textpm2.03}                  & 46.59{\tiny\textpm5.19}                         & 61.91{\tiny\textpm2.45}             & \underline{46.41{\tiny\textpm2.28}} & 63.15{\tiny\textpm2.13}             & 44.53{\tiny\textpm2.12}             \\
        SwinUNETR~\cite{hatamizadeh2022swinunetr} & 6.3~M                            & \textbf{79.97{\tiny\textpm1.13}}     & \underline{26.41{\tiny\textpm3.05}}      & \underline{71.31{\tiny\textpm2.30}}      & \underline{38.96{\tiny\textpm1.04}}             & 64.03{\tiny\textpm0.66}             & 51.38{\tiny\textpm1.48}             & \underline{71.77{\tiny\textpm0.68}} & \underline{38.91{\tiny\textpm1.36}} \\
        TransUNet~\cite{chen2021transunet}        & 3.5~M                            & 76.98{\tiny\textpm2.74}              & 29.93{\tiny\textpm2.71}                  & 57.23{\tiny\textpm2.48}                  & 71.13{\tiny\textpm6.05}                         & 64.88{\tiny\textpm1.60}             & 58.42{\tiny\textpm3.84}             & 66.36{\tiny\textpm0.49}             & 53.16{\tiny\textpm2.17}             \\
        Swin-Unet~\cite{cao2022swinunet}          & 3.4~M                            & 76.47{\tiny\textpm1.26}              & 35.27{\tiny\textpm1.81}                  & 56.55{\tiny\textpm6.75}                  & 67.33{\tiny\textpm4.69}                         & \underline{67.91{\tiny\textpm1.06}} & 59.55{\tiny\textpm4.38}             & 66.98{\tiny\textpm2.79}             & 54.05{\tiny\textpm2.46}             \\
        VM-UNet~\cite{ruan2024vmunet}             & 2.0~M                            & 73.95{\tiny\textpm0.87}              & 29.84{\tiny\textpm2.93}                  & 38.84{\tiny\textpm6.30}                  & 49.37{\tiny\textpm8.44}                         & 51.84{\tiny\textpm5.67}             & 49.30{\tiny\textpm1.22}             & 54.88{\tiny\textpm2.01}             & 42.83{\tiny\textpm3.29}             \\
        \textbf{RIBR}                             & 0.4~M                            & \underline{79.52{\tiny\textpm1.38}}  & \textbf{23.21{\tiny\textpm1.86}}         & \textbf{72.11{\tiny\textpm2.24}}         & \textbf{34.03{\tiny\textpm2.32}}                & \textbf{74.03{\tiny\textpm0.81}}    & \textbf{39.33{\tiny\textpm2.21}}    & \textbf{75.22{\tiny\textpm1.33}}    & \textbf{32.19{\tiny\textpm1.76}}    \\
        \midrule
        \textbf{Method}                           & \textbf{Params}                  & \multicolumn{2}{c|}{\textbf{BUSI}}   & \multicolumn{2}{c|}{\textbf{Breast-USG}} & \multicolumn{2}{c|}{\textbf{BUS-UCLM}}   & \multicolumn{2}{c}{\textbf{Breast Average}}                                                                                                                                                             \\
        \midrule
        U-Net~\cite{ronneberger2015unet}          & 1.8~M                            & 60.07{\tiny\textpm4.07}              & 35.14{\tiny\textpm9.33}                  & 20.53{\tiny\textpm3.78}                  & \underline{42.76{\tiny\textpm6.41}}             & 44.76{\tiny\textpm2.34}             & \underline{43.23{\tiny\textpm1.40}} & 41.79{\tiny\textpm1.59}             & \underline{40.38{\tiny\textpm4.99}} \\
        U-Net++~\cite{zhou2018unetpp}             & 2.3~M                            & 60.60{\tiny\textpm9.49}              & 37.52{\tiny\textpm2.70}                  & 25.06{\tiny\textpm9.54}                  & 44.84{\tiny\textpm5.72}                         & 49.17{\tiny\textpm10.01}            & 56.00{\tiny\textpm17.20}            & 44.94{\tiny\textpm9.41}             & 46.12{\tiny\textpm5.78}             \\
        Attention U-Net~\cite{oktay2018attention} & 2.0~M                            & 62.16{\tiny\textpm6.40}              & 40.30{\tiny\textpm6.82}                  & 29.07{\tiny\textpm9.61}                  & 47.74{\tiny\textpm6.40}                         & 45.54{\tiny\textpm5.01}             & 54.33{\tiny\textpm14.35}            & 45.59{\tiny\textpm5.22}             & 47.46{\tiny\textpm8.21}             \\
        MetaSeg~\cite{vyas2025metaseg}            & 0.1~M                            & 40.49{\tiny\textpm5.71}              & 67.05{\tiny\textpm4.37}                  & 45.96{\tiny\textpm0.59}                  & 61.78{\tiny\textpm2.42}                         & 39.90{\tiny\textpm0.46}             & 80.13{\tiny\textpm1.99}             & 42.12{\tiny\textpm1.69}             & 69.66{\tiny\textpm0.29}             \\
        DynUNet~\cite{isensee2021nnunet}          & 2.0~M                            & 73.25{\tiny\textpm2.11}              & 38.86{\tiny\textpm6.27}                  & 62.23{\tiny\textpm0.52}                  & 57.44{\tiny\textpm1.57}                         & 59.02{\tiny\textpm0.58}             & 76.45{\tiny\textpm6.74}             & 64.83{\tiny\textpm0.61}             & 57.58{\tiny\textpm4.17}             \\
        UNETR~\cite{hatamizadeh2022unetr}         & 13.9~M                           & 69.85{\tiny\textpm3.97}              & 39.38{\tiny\textpm4.37}                  & 61.22{\tiny\textpm1.03}                  & 51.80{\tiny\textpm1.57}                         & 60.50{\tiny\textpm1.69}             & 67.50{\tiny\textpm4.26}             & 63.86{\tiny\textpm1.32}             & 52.89{\tiny\textpm0.76}             \\
        SwinUNETR~\cite{hatamizadeh2022swinunetr} & 6.3~M                            & \textbf{77.02{\tiny\textpm2.93}}     & \underline{29.64{\tiny\textpm7.58}}      & \textbf{64.90{\tiny\textpm1.58}}         & 51.32{\tiny\textpm3.53}                         & \textbf{64.10{\tiny\textpm1.45}}    & 63.38{\tiny\textpm3.85}             & \textbf{68.68{\tiny\textpm0.81}}    & 48.11{\tiny\textpm2.89}             \\
        TransUNet~\cite{chen2021transunet}        & 3.5~M                            & 72.62{\tiny\textpm1.85}              & 37.61{\tiny\textpm5.66}                  & 56.31{\tiny\textpm2.56}                  & 59.06{\tiny\textpm7.36}                         & 55.63{\tiny\textpm1.74}             & 70.02{\tiny\textpm2.30}             & 61.52{\tiny\textpm0.59}             & 55.56{\tiny\textpm1.98}             \\
        Swin-Unet~\cite{cao2022swinunet}          & 3.4~M                            & 70.49{\tiny\textpm2.63}              & 31.56{\tiny\textpm3.96}                  & 48.32{\tiny\textpm8.89}                  & 55.44{\tiny\textpm13.25}                        & 56.47{\tiny\textpm3.04}             & 61.27{\tiny\textpm8.76}             & 58.43{\tiny\textpm3.79}             & 49.42{\tiny\textpm6.14}             \\
        VM-UNet~\cite{ruan2024vmunet}             & 2.0~M                            & 65.85{\tiny\textpm6.14}              & 33.94{\tiny\textpm2.21}                  & 32.68{\tiny\textpm11.37}                 & 46.52{\tiny\textpm3.22}                         & 50.03{\tiny\textpm8.71}             & 48.36{\tiny\textpm8.55}             & 49.52{\tiny\textpm8.61}             & 42.94{\tiny\textpm4.27}             \\
        \textbf{RIBR}                             & 0.4~M                            & \underline{74.56{\tiny\textpm4.62}}  & \textbf{28.13{\tiny\textpm6.64}}         & \underline{64.72{\tiny\textpm2.37}}      & \textbf{39.15{\tiny\textpm3.02}}                & \underline{62.54{\tiny\textpm2.73}} & \textbf{42.72{\tiny\textpm4.41}}    & \underline{67.27{\tiny\textpm2.20}} & \textbf{36.67{\tiny\textpm3.26}}    \\
        \midrule
        \textbf{Method}                           & \textbf{Params}                  & \multicolumn{2}{c|}{\textbf{DDTI}}   & \multicolumn{2}{c|}{\textbf{TN3K}}       & \multicolumn{2}{c|}{\textbf{Prostate}}   & \multicolumn{2}{c}{\textbf{All Average}}                                                                                                                                                                \\
        \midrule
        U-Net~\cite{ronneberger2015unet}          & 1.8~M                            & 86.19{\tiny\textpm0.95}              & 25.76{\tiny\textpm1.83}                  & 80.28{\tiny\textpm0.51}                  & 29.21{\tiny\textpm1.16}                         & 89.64{\tiny\textpm0.65}             & 21.95{\tiny\textpm1.10}             & 62.15{\tiny\textpm1.89}             & 36.74{\tiny\textpm2.96}             \\
        U-Net++~\cite{zhou2018unetpp}             & 2.3~M                            & 87.01{\tiny\textpm0.88}              & 24.24{\tiny\textpm0.69}                  & 80.35{\tiny\textpm0.15}                  & 27.66{\tiny\textpm1.83}                         & 89.57{\tiny\textpm1.04}             & 23.39{\tiny\textpm1.25}             & 63.60{\tiny\textpm3.33}             & 39.90{\tiny\textpm3.26}             \\
        Attention U-Net~\cite{oktay2018attention} & 2.0~M                            & 83.85{\tiny\textpm0.76}              & 27.47{\tiny\textpm2.07}                  & 80.08{\tiny\textpm0.48}                  & 29.22{\tiny\textpm0.86}                         & 89.00{\tiny\textpm1.97}             & 24.70{\tiny\textpm2.32}             & 62.74{\tiny\textpm1.71}             & 40.17{\tiny\textpm2.47}             \\
        MetaSeg~\cite{vyas2025metaseg}            & 0.1~M                            & 79.36{\tiny\textpm0.72}              & 32.05{\tiny\textpm1.30}                  & 39.24{\tiny\textpm0.91}                  & 65.18{\tiny\textpm1.94}                         & 72.90{\tiny\textpm0.37}             & 59.87{\tiny\textpm1.07}             & 49.20{\tiny\textpm0.23}             & 57.84{\tiny\textpm0.39}             \\
        DynUNet~\cite{isensee2021nnunet}          & 2.0~M                            & 83.92{\tiny\textpm0.63}              & 31.74{\tiny\textpm1.27}                  & 79.41{\tiny\textpm0.43}                  & 31.09{\tiny\textpm0.88}                         & 90.87{\tiny\textpm0.07}             & 27.29{\tiny\textpm0.11}             & 70.02{\tiny\textpm1.19}             & 46.48{\tiny\textpm0.17}             \\
        UNETR~\cite{hatamizadeh2022unetr}         & 13.9~M                           & 85.96{\tiny\textpm0.24}              & \underline{22.97{\tiny\textpm0.33}}      & 70.49{\tiny\textpm0.32}                  & 39.84{\tiny\textpm0.70}                         & 89.84{\tiny\textpm0.16}             & 25.56{\tiny\textpm0.93}             & 69.70{\tiny\textpm0.40}             & 42.30{\tiny\textpm0.84}             \\
        SwinUNETR~\cite{hatamizadeh2022swinunetr} & 6.3~M                            & \underline{89.04{\tiny\textpm0.22}}  & 23.02{\tiny\textpm1.88}                  & 79.99{\tiny\textpm0.29}                  & 28.20{\tiny\textpm0.67}                         & \underline{92.11{\tiny\textpm0.14}} & \underline{17.82{\tiny\textpm0.21}} & \underline{75.83{\tiny\textpm0.45}} & 36.68{\tiny\textpm0.85}             \\
        TransUNet~\cite{chen2021transunet}        & 3.5~M                            & 85.41{\tiny\textpm0.67}              & 27.41{\tiny\textpm1.37}                  & \textbf{82.17{\tiny\textpm0.28}}         & \underline{25.24{\tiny\textpm0.59}}             & 91.63{\tiny\textpm0.06}             & 18.72{\tiny\textpm0.26}             & 71.43{\tiny\textpm0.18}             & 44.17{\tiny\textpm1.02}             \\
        Swin-Unet~\cite{cao2022swinunet}          & 3.4~M                            & 85.71{\tiny\textpm1.37}              & 26.63{\tiny\textpm3.16}                  & \underline{81.44{\tiny\textpm0.55}}      & 26.54{\tiny\textpm2.42}                         & 91.89{\tiny\textpm0.39}             & 18.60{\tiny\textpm0.52}             & 70.58{\tiny\textpm0.27}             & 42.46{\tiny\textpm3.19}             \\
        VM-UNet~\cite{ruan2024vmunet}             & 2.0~M                            & 86.43{\tiny\textpm0.50}              & 25.26{\tiny\textpm0.32}                  & 80.96{\tiny\textpm0.57}                  & 26.41{\tiny\textpm1.99}                         & 89.79{\tiny\textpm0.89}             & 20.44{\tiny\textpm1.99}             & 63.37{\tiny\textpm3.22}             & \underline{36.60{\tiny\textpm1.15}} \\
        \textbf{RIBR}                             & 0.4~M                            & \textbf{89.23{\tiny\textpm0.53}}     & \textbf{20.73{\tiny\textpm1.04}}         & 80.86{\tiny\textpm0.53}                  & \textbf{22.85{\tiny\textpm1.02}}                & \textbf{92.42{\tiny\textpm0.27}}    & \textbf{15.79{\tiny\textpm0.64}}    & \textbf{76.67{\tiny\textpm0.42}}    & \textbf{29.55{\tiny\textpm0.55}}    \\
        \bottomrule
    \end{tabular}
\end{table*}

\begin{figure*}[t]
    \centering
    \includegraphics[width=0.85\textwidth]{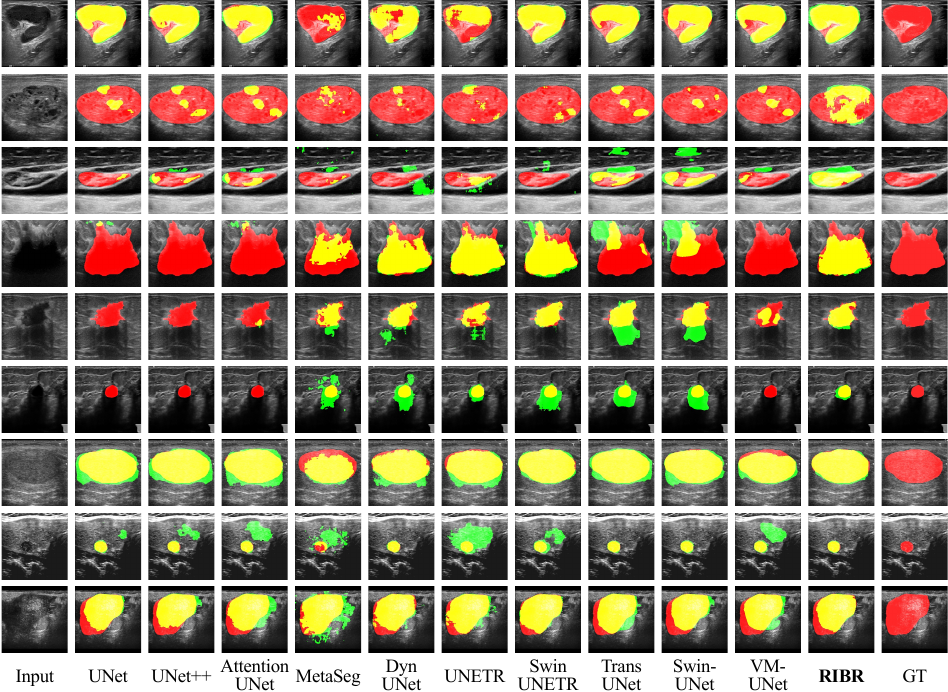}
    \caption{Qualitative comparison. Rows show representative cases with input, baselines, RIBR, and ground truth (GT). Red, green, and yellow indicate ground truth, false positives, and true positives, respectively.}
    \label{fig:comparison}
\end{figure*}

All models use full supervision, shared splits, seeds 1--3, and up to 120 epochs with early stopping. DynUNet uses a fixed MONAI configuration. RIBR and its ablations use random initialization, single-channel $256 \times 256$ inputs in $[0,1]$, and AdamW with constant learning rate $1.5\times10^{-4}$, weight decay 0.01, and batch size 1. Validation runs every two epochs, stopping after 20 non-improving checks; checkpoints maximize $\mathrm{Dice}+\mathrm{IoU}-\mathrm{HD95}/256-\mathrm{ASD}/64$, using Dice and IoU in $[0,1]$ and distances in pixels.

BRI uses $72\rightarrow32\rightarrow32\rightarrow2$ layers and $\beta=1$. GSBR truncates distances at 16 pixels and applies distance supervision less than four pixels from the contour. Loss weights are $(\lambda_f,\lambda_d,\lambda_c)=(1,1,0.5)$ and $(\lambda_b,\lambda_s,\lambda_p,\lambda_r)=(0.6,0.18,0.03,0.06)$; $\kappa=5$ and $(\tau,\gamma,\lambda)=(0.65,12,0.25)$. Ablations retain these settings for active modules.

\subsection{Evaluation Metrics}
\label{sec:metrics}
We report foreground Dice (\%) and HD95, the maximum of the two directed 95th-percentile surface distances. Ablations also report intersection over union (IoU) and average surface distance (ASD), measured from prediction to reference.  For each seed, image scores are averaged within datasets, then equally across the three LN datasets, three breast datasets, or all nine datasets; ablations use the same LN macro. Results are mean $\pm$ sample standard deviation (SD) over three seeds.

\subsection{Main Results}
RIBR achieves the best overall macro-average among evaluated configurations in~\Cref{tab:results}, with $76.67{\pm}0.42\%$ Dice and $29.55{\pm}0.55$ HD95 in pixels. SwinUNETR is the closest Dice competitor at $75.83{\pm}0.45\%$; VM-UNet gives the next-lowest HD95, $36.60{\pm}1.15$ pixels. RIBR achieves the lowest HD95 on all nine datasets.

On external LymphUS-C1/C2, RIBR leads both metrics. Its LN macro is $75.22{\pm}1.33\%$ Dice and $32.19{\pm}1.76$ HD95, improving over SwinUNETR by 3.45 Dice points and 6.72 pixels. SwinUNETR leads Dice on LN-INT and all three breast datasets; RIBR ranks second in breast macro Dice at $67.27{\pm}2.20\%$, with the lowest macro HD95 of $36.67{\pm}3.26$ pixels. RIBR leads both metrics on DDTI and Prostate, but ranks fourth in TN3K Dice. Its advantage is therefore more consistent in boundary distance than overlap.

\Cref{fig:comparison} shows representative contour erosion, leakage, and fragmentation in baseline predictions. RIBR generally follows the reference boundary more closely while preserving the main region, consistent with its HD95 results.

\subsection{Ablation and Efficiency}
\label{sec:ablation}
\begin{table}[htb]
    \caption{LN module ablation, dataset macro mean $\pm$ SD over three seeds. Distances are in pixels. Base is the convolutional predictor.}
    \label{tab:ablation_modules}
    \centering
    \setlength{\tabcolsep}{2.8pt}
    \begin{tabular}{c|cccc|rrrr}
        \toprule
        Exp. & Base       & BRI        & GSBR       & RRC        & Dice\%$\uparrow$                 & IoU\%$\uparrow$                  & HD95$\downarrow$                 & ASD$\downarrow$                  \\
        \midrule
        (a)  & \checkmark &            &            &            & 72.41{\tiny\textpm1.28}          & 61.23{\tiny\textpm1.34}          & 37.49{\tiny\textpm3.62}          & 13.28{\tiny\textpm1.36}          \\
        (b)  & \checkmark &            & \checkmark &            & 72.22{\tiny\textpm2.89}          & 61.04{\tiny\textpm3.32}          & 35.74{\tiny\textpm3.61}          & 12.40{\tiny\textpm1.41}          \\
        (c)  & \checkmark & \checkmark &            &            & 73.67{\tiny\textpm2.01}          & 62.79{\tiny\textpm2.09}          & 35.91{\tiny\textpm1.30}          & 12.60{\tiny\textpm0.66}          \\
        (d)  & \checkmark & \checkmark & \checkmark &            & 72.49{\tiny\textpm0.72}          & 61.41{\tiny\textpm1.01}          & 38.38{\tiny\textpm3.12}          & 13.72{\tiny\textpm1.87}          \\
        (e)  & \checkmark & \checkmark &            & \checkmark & 73.03{\tiny\textpm2.05}          & 61.92{\tiny\textpm2.67}          & 36.13{\tiny\textpm4.18}          & 12.95{\tiny\textpm1.75}          \\
        (f)  & \checkmark & \checkmark & \checkmark & \checkmark & \textbf{75.22{\tiny\textpm1.33}} & \textbf{64.53{\tiny\textpm1.50}} & \textbf{32.19{\tiny\textpm1.76}} & \textbf{11.40{\tiny\textpm1.06}} \\
        \bottomrule
    \end{tabular}
\end{table}

In~\Cref{tab:ablation_modules}, Full RIBR improves LN macro Dice from $72.41{\pm}1.28\%$ to $75.22{\pm}1.33\%$ and reduces HD95 from $37.49{\pm}3.62$ to $32.19{\pm}1.76$ pixels relative to Conv-only. BRI alone improves both metrics; GSBR alone reduces distance with slightly lower Dice. BRI+GSBR has higher surface errors than BRI-only, and BRI+RRC also falls short of Full. The complete configuration gives the best values in all four reported metrics for this LN comparison.

\begin{table}[htb]
	\caption{Model forward-pass cost.}
	\label{tab:runtime_cost}
	\centering
	\begin{tabular}{lrrrr}
		\toprule
		Method    & Params (M) & GFLOPs & Latency (ms) & FPS    \\
		\midrule
		U-Net     & 1.81       & 5.88   & 1.38 & 726.76 \\
		SwinUNETR & 6.31       & 7.38   & 9.71 & 102.96 \\
		RIBR      & 0.42       & 30.26  & 4.23 & 236.57 \\
		\bottomrule
	\end{tabular}
\end{table}

\Cref{tab:runtime_cost} reports model forward-pass cost at 32-bit floating-point precision and batch size 1 on an RTX 5090, averaged over 300 runs after 50 warm-ups. Floating-point operations (FLOPs) count convolutional/linear multiply-adds as two operations. RIBR uses 0.42 M parameters and takes 4.23 ms per forward, between U-Net and SwinUNETR. Its larger FLOP count reflects dense implicit computation at all 65,536 pixels.

\subsection{Discussion}
External LN and breast cohorts assess performance under acquisition and center shifts; thyroid and prostate experiments assess segmentation on separately trained tasks. The evaluation is retrospective and binary. Confident errors can suppress useful corrections, and uncertain artifacts can receive larger gates. Baseline protocol differences restrict architectural comparisons; resized-grid distances have limited physical interpretation.

\section{Conclusion}
RIBR combines a compact convolutional predictor with risk-routed implicit residuals and contour-sensitive training. Across nine US datasets, it achieves the best overall macro-average among evaluated configurations and the lowest HD95 on every dataset. Dice rankings vary by task, and LN ablations support the complete configuration for boundary-sensitive segmentation.

\bibliographystyle{IEEEtran}
\bibliography{references}

\end{document}